\documentclass{article}

\usepackage{arxiv}

\usepackage[utf8]{inputenc} 
\usepackage[T1]{fontenc}    
\usepackage{hyperref}       
\usepackage{url}            
\usepackage{booktabs}       
\usepackage{amsfonts}       
\usepackage{nicefrac}       
\usepackage{microtype}      
\usepackage{lipsum}		
\usepackage{graphicx}
\usepackage{natbib}
\usepackage{doi}

\usepackage{amsmath}
\usepackage[ruled]{algorithm2e}
\usepackage{mathtools}
\usepackage{booktabs,makecell,multirow}
\usepackage{algorithmic}

\usepackage{balance} 
\usepackage{color}
\definecolor{mygreen}{RGB}{70,102,94}
\definecolor{myyellow}{RGB}{234,178,0}
\definecolor{myper}{RGB}{128,82,110}
\definecolor{myred}{RGB}{192,0,0}
\definecolor{myblue}{RGB}{216,223,224}
\definecolor{mygreenshade}{RGB}{235,243,228}
\definecolor{myredshade}{RGB}{250,237,233}
\definecolor{myyellowshade}{RGB}{254,248,232}
\usepackage{bbm}
\usepackage{textcomp}
\usepackage{amssymb}

\title{Causal Prompting Model-based Offline Reinforcement Learning}

\author{ 
    Xuehui Yu\\
    Faculty of Computing\\ Harbin Institute of Technology\\ Harbin, Heilongjiang, China\\
	\texttt{yuxuehui@stu.hit.edu.cn} \\
    \And
    Yi Guan \\
	Faculty of Computing\\ Harbin Institute of Technology\\ Harbin, Heilongjiang, China\\
	\texttt{guanyi@hit.edu.cn}\\
    \And
    Rujia Shen \\
	Faculty of Computing\\ Harbin Institute of Technology\\ Harbin, Heilongjiang, China\\
	\texttt{shenrujia@stu.hit.edu.cn} \\
	\And
    Xin Li \\
	Faculty of Computing\\ Harbin Institute of Technology\\ Harbin, Heilongjiang, China\\
	\texttt{22S103169@stu.hit.edu.cn} \\
	\And
    Chen Tang \\
	University of Surrey\\ Guildford, England, United Kingdom\\
	\texttt{travistang@foxmail.com} \\
	\And
    Jingchi Jiang\\
    The Artificial Intelligence Institute\\ Harbin Institute of Technology\\ Harbin, Heilongjiang, China\\
	\texttt{jiangjingchi@hit.edu.cn} \\
}

\renewcommand{\shorttitle}{\textit{arXiv} Template}

\hypersetup{
pdftitle={A template for the arxiv style},
pdfsubject={q-bio.NC, q-bio.QM},
pdfauthor={David S.~Hippocampus, Elias D.~Striatum},
pdfkeywords={First keyword, Second keyword, More},
}

\begin{document}
\maketitle

\begin{abstract}
	Model-based offline Reinforcement Learning (RL) allows agents to fully utilise pre-collected datasets without requiring additional or unethical explorations. However, applying model-based offline RL to online systems presents challenges, primarily due to the highly suboptimal (noise-filled) and diverse nature of datasets generated by online systems. To tackle these issues, we introduce the Causal Prompting Reinforcement Learning (CPRL) framework, designed for highly suboptimal and resource-constrained online scenarios. The initial phase of CPRL involves the introduction of the Hidden-Parameter Block Causal Prompting Dynamic (Hip-BCPD) to model environmental dynamics. This approach utilises invariant causal prompts and aligns hidden parameters to generalise to new and diverse online users. In the subsequent phase, a single policy is trained to address multiple tasks through the amalgamation of reusable skills, circumventing the need for training from scratch. Experiments conducted across datasets with varying levels of noise, including simulation-based and real-world offline datasets from the \textit{Dnurse} APP, demonstrate that our proposed method can make robust decisions in out-of-distribution and noisy environments, outperforming contemporary algorithms. Additionally, we separately verify the contributions of Hip-BCPDs and the skill-reuse strategy to the robustness of performance. We further analyse the visualised structure of Hip-BCPD and the interpretability of sub-skills. We released our source code and the first ever real-world medical dataset for precise medical decision-making tasks.
\end{abstract}

\keywords{Generalisation \and Causal Reinforcement Learning \and Reinforcement Learning \and Prompting}

\section{Introduction}\label{sec:introduction}
Despite the demonstrated efficacy of Reinforcement Learning (RL) methodologies across various domains \citep{schulman2017proximal,kaiser2019model,tang2022recent,rolf2023review}, e.g., CartPole and Atari, there remains a notable scarcity of endeavours directed towards the application of RL techniques within real-world medical online systems. One of the primary challenges impeding such endeavours pertains to the high safety requirement of medical practitioners. It is very dangerous to directly use an online system in exploratory interactions between patients and the medical online systems\citep{vouros2022explainable,bostrom2003ethical,amin2017repeated}. To address this challenge, offline RL \citep{levine2020offline} is developed to train RL agents only with pre-collected offline datasets and thus can avoid the danger of online explorations. Typically, directly applying pre-trained offline RL agents on real-world online systems suffers from model bias brought by the noises contained in online systems. This phenomenon is primarily attributed to the individual inaccuracies of data, especially those consisting of users' self-reported lifestyle data in healthcare systems \citep{sarker2021machine}. In addition, the features of datasets vary among different populations, which is also called distribution shifts \citep{al2016impact, limayem2003force}. Therefore, in this paper, we aim to propose a novel offline RL framework, which provides medical advice to diverse populations based on suboptimal (containing noise or even errors) pre-collected dataset from an online medical system\footnote{\textit{Dnurse}: https://www.dnurse.com/v2/en/ \label{dnurse_web}}.

Up to date, there have been limited exploration in offline RL studies that specifically address the challenges posed by suboptimal datasets \citep{liu2024design,liu2019off,wang2020statistical,zolna2020offline}. These studies encounter various obstacles when applied to our task. Broadly, existing research in this domain can be categorised into two distinct directions. One strand of research \citep{fujimoto2019off,kumar2019stabilizing,kumar2020conservative,yu2020mopo} endeavours to evaluate RL approaches using simulated suboptimal data. However, the efficacy of these approaches remains unproven when confronted with real-world suboptimal data characterised by inherently noisy distributions that are more intricate and less predictable. Conversely, another group of studies \citep{zhou2023real,xu2021positive,konyushkova2020semi,zolna2020offline,wang2023mimicplay} primarily focuses on preprocessing suboptimal data rather than proposing a robust RL framework to contend with noise. For example, learnable reward models \citep{xu2021positive,konyushkova2020semi,zolna2020offline} are proposed to relabel suboptimal data with low rewards, guiding RL agents to avoid noise and errors. However, additional learnable parameters bring more complex calculations leading to less robustness. In addition, existing RL approaches struggle to deal with data collected from diverse populations. \citep{wang2023mimicplay} trained a single policy across tasks require large in-domain perfectly aligned human-robot data and are not capable of leveraging passive web data. For generalisation across all tasks during testing, offline datasets must encompass a diverse range of states, necessitating substantial data volume that may not be feasible in resource-limited settings. In contrast to these approaches, our study explores offline RL agents trained without complex computations across multiple tasks and under resource-limited scenarios. In contrast to these approaches, our study explores offline RL agents trained without complex computations across multiple tasks and under resource-limited scenarios.

\begin{figure*}[!ht]
    \centering
    \includegraphics[width=1\linewidth]{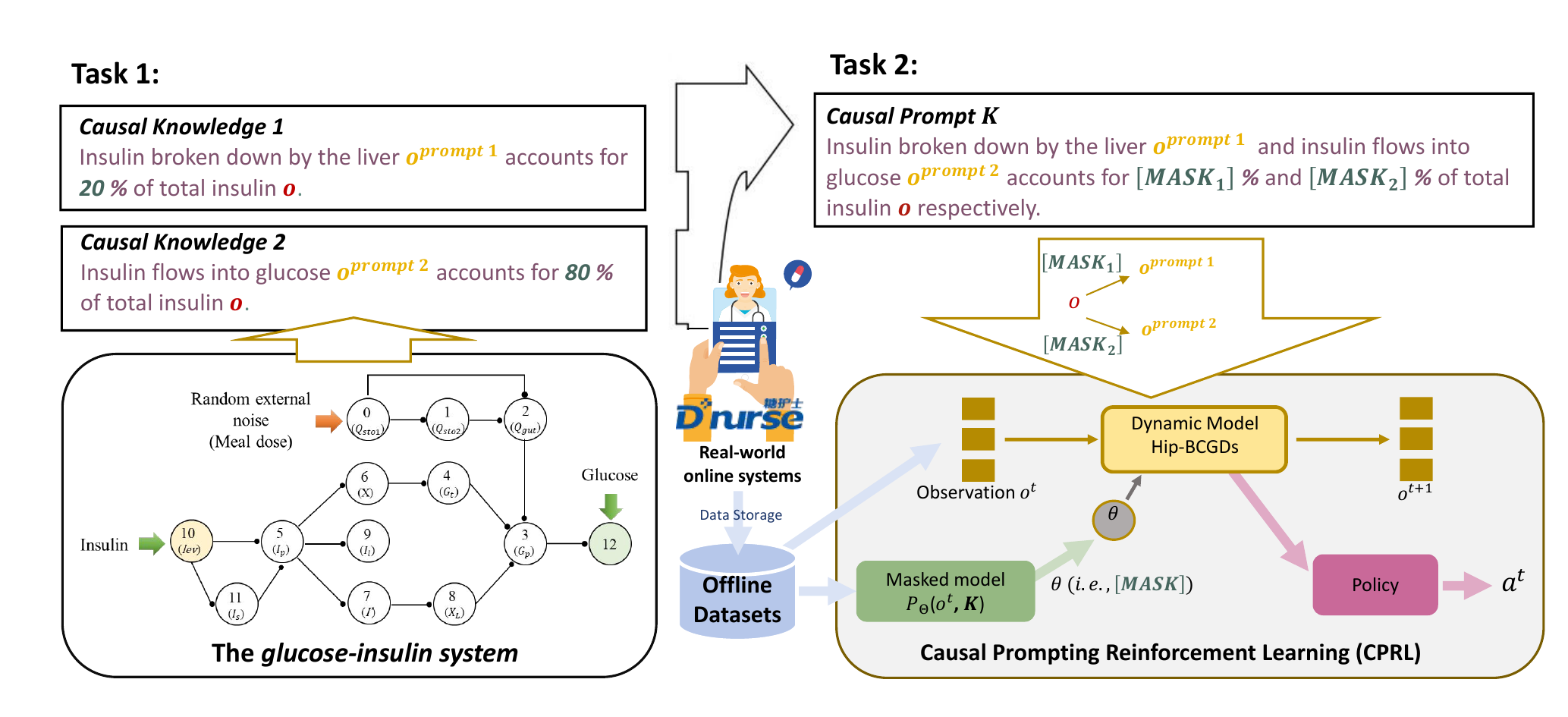}
    \caption{Illustration of the Causal Prompting Reinforcement Learning (CPRL) framework. CPRL learns on suboptimal offline datasets using the Causal Prompt $\mathcal{K}$ as a guiding mechanism. \textbf{\textit{Causal prompt}} $\mathcal{K}$ leverages causal knowledge (on the top left) as a template, amalgamating the original input $o$ with reconstructed observations $o^{prompt}$. At the bottom left of the figure, the \textit{glucose-insulin system} represents the pre-trained model, while the graphical structure visualises causal knowledge. \textit{Masked model} $P_\Theta$ (green box) generates task-specific hidden parameters $\theta \in \Theta$ (i.e., $[MASK]$ in the causal prompt $\mathcal{K}$). The \textbf{\textit{CPRL}} framework predominantly consists of two processes: 1) learning dynamic models (grey box); and 2) learning policies (pink box). Agents acquire dynamic models from offline datasets with the guidance of the causal prompt $\mathcal{K}$ and hidden parameters $\theta$. The constructed dynamic model is subsequently utilised for downstream policy learning.}
    \label{fig1}
\end{figure*}

To achieve the aforementioned objectives, our research draws inspiration from the theory of \textit{causal invariance hypothesis} \citep{wang2021task,volodin2020resolving,seitzer2021causal,gasse2021causal}, which posits that the data generation mechanism remains invariant (or partially invariant) across multiple tasks. Consequently, we aim to build a multi-task offline RL framework wherein dynamic models share common data generation mechanisms across various tasks. We construct dynamic models grounded in invariant causal knowledge, denoting the relationship where a single cause produces a consistent effect across diverse tasks \citep{park2022causal}.
Incorporating invariant causal relationships towards reducing mitigating underspecification and overfitting is increasingly gaining traction \citep{kyono2020castle,creager2021environment,lin2022zin,chen2023rethinking}, largely due to their implications for causality. Examples can be different devices for capturing the images, or the hospitals at which the patient data are collected. Broadly, it can be considered a set of conditions, interpreted as the `context' of the data. As an important indication of distinguishing causality from suboptimal datasets, the causal relationship should be invariant across all possible tasks/environments.
Via this approach, our proposed \textbf{Causal Prompting Reinforcement Learning (CPRL)}, a novel model-based offline RL framework integrating prompt learning via causal invariance, mitigate the challenge of learning from extensive and heterogeneous datasets. Similar to conventional multi-task RL \citep{hessel2019multi, wilson2007multi, yang2020multi}, our framework comprises two primary processes: (1) learning dynamic models with shared invariant causal promptings, and (2) learning a policy with invariant behavioural strategies. Specifically,
\begin{itemize}
\item [(1)]
To mitigate the risk of overfitting to suboptimal data, we propose the Hidden-Parameter Block Causal Prompting Dynamic (Hip-BCPD) method, which leverages invariant causal promptings when adapting to diverse environments. The hidden parameters within the Hip-BCPD are trained to align with the underlying environmental settings, thereby facilitating model generalisation across diverse environments present in the suboptimal data. Causal prompts, serving as contextual cues during the dynamic model adaptation, are derived from the output of a pre-trained language model. This adaptation process models diverse environments without introducing additional learnable parameters or increasing training complexity \citep{xin2022rethinking, wang2022learning}, resulting in more robust features resilient to noise in the suboptimal data.
\item[(2)] In behaviour learning, we use a Causal Coupled Mechanism (CCM) \citep{yu2022causal}, a policy with a skill-reuse strategy. By acquiring reusable skills, this strategy enables a single policy to harness these shared skills to learn to solve multiple tasks quickly. 
\end{itemize}

\textbf{Summary of Contributions:}
\begin{itemize}
\item We propose CPRL, a framework that enables offline learning from a large set of diverse and suboptimal datasets under the constraint of causal prompts.
\item We propose a novel dynamic modelling method, namely Hip-BCPDs, aimed at mitigating overfitting to suboptimal data via restoring invisible environmental variables with the prompt of causal knowledge.
\item We introduce and release a novel real-world medical dataset for the task of precise medical decisions. To our knowledge, this is the first attempt to apply an offline RL approach on a real-world medical dataset, rather than relying on simulated data.
\item We conducted a range of experiments to demonstrate the effectiveness of CPRL in large-scale suboptimality offline data and confirm that CPRL substantially outperforms other state-of-the-art RL baselines.
\end{itemize}

\section{Background}

In this section, we introduce the relevant theoretical background knowledge for learning the dynamic models and learning policies, respectively.

\begin{figure*}[!h]
    \centering
    \includegraphics[width=1\linewidth]{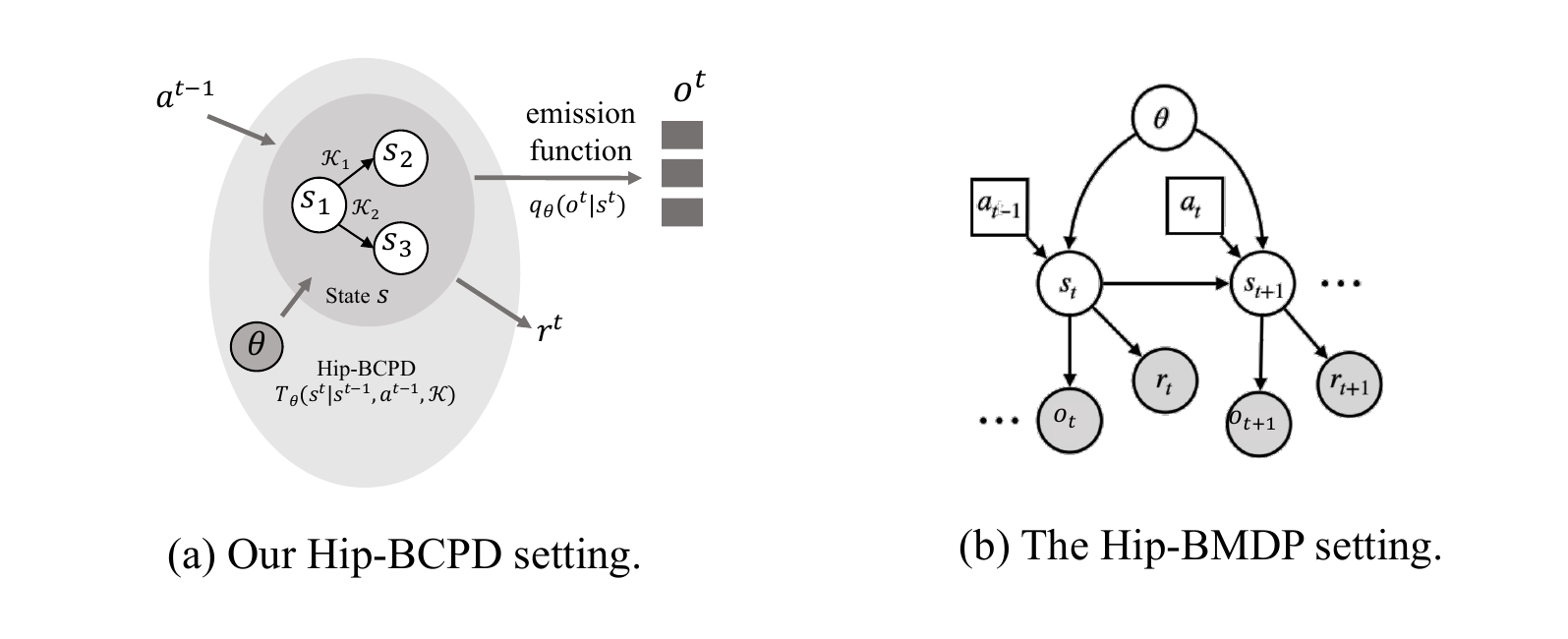}
    \caption{Visualizations of Hip-BCPD and Hip-BMDP settings. In the Hip-BCPD setting, causal promptings $\mathcal{K}_1$ and $\mathcal{K}_2$ act as edges connecting variables in state $s=[s_1,s_2,s_3]$ into a causal graph. Instead, the state $s$ is latent and without a graph structure in the HiP-BMDP setting.
    In Hip-BMDP and Hip-BCPD, hidden parameters $\theta$ decide transition distributions $T_{\theta}$ and emission function $q$ vary among diverse environments. Figure \ref{fig2}(b) is cited from \citep{zhang2020learning}.}
    \label{fig2}
\end{figure*}

In the dynamic modelling part, we first introduce the base environment dynamics - the \textbf{Hidden-Parameter Block Markov Decision Process (Hip-BMDP)} \citep{zhang2020learning}. The Hip-BMDP family, graphically presented in Figure \ref{fig2} (b), is described by tuple $\left \langle \mathcal{S},\mathcal{A},\mathcal{O},\mathcal{R},\Theta,P_\Theta,q_{\theta},T_\theta,\gamma \right \rangle$, where $\mathcal{S}$ is the set of states, $\mathcal{A}$ is the set of actions, $\mathcal{O}$ is the set of observations, $\mathcal{R}:\mathcal{S} \times \mathcal{A} \to \mathbb{R}$ describes the reward function, $\Theta$ is the hidden-parameter space, $T_{\theta}$ describes dynamic of a environment with a hidden parameter $\theta \backsim P_{\Theta}$. The emission function $q_{\theta}(o|s)$ maps states $s \in \mathcal{S}$ to observations $o \in \mathcal{O}$, operating under a one-to-many problem. This indicates that while the state remains invariant, the observations perceived by the agent may have considerable variability. Consequently, the variation in the hidden parameter $\theta$ introduces diversity across tasks in the Hip-BMDP. A pertinent illustration of this phenomenon can be observed in healthcare applications, where data characteristics differ across different populations. The Emission function $q_{\theta}(o|s)$ is based on the \textbf{Block structure} assumption illustrated as follows.

\textbf{Assumption 1} (Block structure \citep{du2019provably}). \textit{Each observation $o$ is generated by a unique state $s$. That is, several disjoint blocks $\mathcal{O}_{s}$, each of them supported by a conditional distribution $q_{\theta}(\cdot|s)$, combine into the observation space $\mathcal{O}$.}

Assumption 1 gives the Markov property in observation space $\mathcal{O}$, different from the partially observable Markov decision process \citep{yarats2021improving}. The assumption of a block structure theory posits that states can be reconstructed from observations. However, it is impossible to reconstruct a complete state; it is more practical to reconstruct states necessary for downstream tasks. The method of reconstructing or representing states by utilising causal structures, a clever approach to transferring shared knowledge across multiple tasks, is gradually attracting the attention of researchers \citep{wang2021task,volodin2020resolving,seitzer2021causal,gasse2021causal}. Inspired by their work, we also employ causal structures to establish dynamic models and reconstruct states. A prerequisite for this is the object-level or event-level abstraction of the observation, which is unavailable in most tasks \citep{abel2022theory,shanahan2022abstraction}. We avoid object-level or event-level abstraction and instead use prompting methods to construct causal relationships. Hence, a base dynamic model -- causal graph dynamic and prompting learning are briefly introduced in the following content.

The dynamic learning problem setting can be approached from an alternative perspective wherein it can be construed as a prediction problem with a target $o$ and predicted observations $o^{t+1}$ in multiple environments, i.e., learning the hidden parameter $\theta$ which includes learnable parameter in $T_\theta$ and $q_{\theta}(o|s)$.
First, we assume that there is an underlying, causal representation of the states $s$ whose constituents are causes of the prediction target $o$, and they interact with each other through a causal graph. This representation $s$ and the transformation $o = g_{causal}(s) = q_{\theta}(\cdot|s)$ are the central components essential for tackling the dynamic learning problem. Further, in a dynamic model, the change of a particular part will affect the interaction between neighbours, thereby inducing global changes. \textbf{Causal Graph Dynamics (CGDs)} define the above global changes as follows:

\textbf{Definition 1} (Causal Graph Dynamics \citep{arrighi2012causal}): \textit{A dynamic $T:\mathcal{G}_{\Sigma,\Delta} \to \mathcal{G}_{\Sigma,\Delta}$ is causal if and only if there exists a radius $r$ and a bound $b$, such that the following two conditions are met:} 
\begin{itemize}
    \item [i)] \textit{Uniform continuity.}
    \begin{align}
        &\forall v',v \in a(v'), \forall G, H \in \mathcal{G}_{\Sigma,\Delta}, \left[ G_v^r = H_v^r \circ T(G)_{v'} = T(H)_{v'} \right] ;
    \end{align}
    \item [ii)] \textit{Boundedness.}
    \begin{equation}
        \forall G \in \mathcal{G}_{\Sigma,\Delta}, \forall v \in G, |\{ v'\in T(G)|v \in a(v') \}| \leq	b,
        \label{eq4.2}
    \end{equation}
\end{itemize}
where $\mathcal{G}_{\Sigma,\Delta}$ is a set of causal graphs. The graph $H$ is the conjugacy of graph $G$ through dynamic $T$, i.e., for all $G \in \mathcal{G}_{\Sigma,\Delta}$, $T(GH) \circ G'T(H)$ and $G'$ is an isomorphism of $G$. $a(\cdot)$ is the antecedent co-dynamics of $T$. CGDs define causal dynamics with time-varying neighbourhoods, which model dynamics caused by neighbour-to-neighbour interactions and time-varying neighbourhoods. Primarily when the agent interacts with the environment, due to the existence of CGD, minor changes in actions may cause dramatic changes of the observed variables. 
In the language of CGDs, this means that the conditional distribution $P(o|s_{Pa_o})$ remains invariant under any interventions on $s$. This principle is also referred to as autonomy \citep{aldrich1989autonomy}, modularity \citep{pearl2009causality} and independence of cause and mechanism \citep{peters2017elements}. This connection brings the model a good generalisation ability on different datasets and reduces errors.

To incorporate causal invariance, we draw inspiration from \textbf{Prompt Learning}, which also effectively utilises the invariant characteristics present in the machine learning process via learning on the causal templates. Prompt learning is the technique of making better use of the knowledge from the pre-trained model. Utilising fill-in-the-blanks (e.g., cloze-style) language prompts as a mechanism to activate the extensive knowledge encapsulated within pre-trained language models, such as BERT \citep{devlin2018bert}, the technique of prompt learning demonstrates notable efficacy across various natural language processing tasks \citep{shin2020autoprompt}. For example, in the entity typing task, with each sentence $o$ containing a marked entity mention $[MASK]$, we aim at predicting the entity type $y\in \mathcal{Y}$ by distinguishing the original input $o_{\text{input}}$ from the prompt $o_{\text{prompt}}$. We use the template $\mathcal{K}$ mapping $o_{\text{input}}$ to $o_{\text{prompt}}$, wherein the model is trained to select the tokens that should be filled in $[MASK]$ under the prompt by maximising $p([MASK]|\mathcal{K})$. Consequently, with the help of prompt learning, the aforementioned problem can be transformed into a masked language modelling problem,
\begin{equation}
p(y\in \mathcal{Y}|o_{\text{input}})=p([MASK]=w\in V_y|\mathcal{K}(o_{\text{input}})).
\end{equation}

When modelling environment dynamics, prompt learning utilises pre-trained models to assist the learning processes of dynamic models.

To facilitate policy learning within complex transition dynamics, we propose a \textbf{Hierarchical Reinforcement Learning} framework. This approach decomposes a complex problem into several sub-problems, aligning well with the boundedness and uniform continuity properties of CGDs. Moreover, the hierarchical structure facilitates knowledge transfer across different problems, allowing component solutions to be efficiently reused in tackling new and increasingly complex challenges. Under the hierarchical framework, the transition dynamic is formulated as a tuple $\left \langle \mathcal{S},\mathcal{A},\mathcal{R}_{a^{high},\Omega}, T_{a^{high}} \right \rangle$, where $T_{a^{high}}(s_{t+1}|s_t) = p(s_{t+1}|s_t, a^{high}) \prod_{i=1}^{K-1}p(\omega_{i+1}|\Omega_i)$. $\Omega$ is a transition function space to describe $K$ stages transiting inside the high-level action $a^{\text{high}}$. Each $\omega$ is the low-level action of $a^{\text{high}}$. All of $\omega$ are relevant to each other. For downstream evaluation of the dynamic, we use \textbf{Causal Coupled Mechanisms (CCMs)} \citep{yu2022causal}, a hierarchical RL framework with a skill-reuse strategy. In the hierarchical framework of CCMs, the high-level policy divides a CGD into $K$ sub-CGDs and generates a strategy to reuse optimal skills automatically, and the low-level policy learns the skill of controlling a sub-CGD stably.


\section{Causal Prompting Reinforcement Learning}

In real-world applications, decision-support agents often need to deal with suboptimal and diverse large-scale offline datasets and face the challenges of generalising to a new online user. The CPRL framework trains agents to be generalisable to diverse users while demonstrating strong robustness against highly suboptimal datasets. This advantage stems from the utilisation of its dynamic model - Hip-BCPD, coupled with its hierarchical policy architecture. Specifically, 1) in Section \ref{3.1}, we introduce how to use causal prompting to learn the Hip-BCPD model to enhance its robustness, and the design of hidden parameters enables it to encode different users' features for effective generalisation; 2) Built upon the understanding of the world through Hip-BCPD, agents formulate actions through a hierarchical policy (introduced in Section \ref{3.2}). Instead of solving each task individually from scratch, hierarchical policy architecture can leverage the fact that hidden parameters of variation often relate to different environments to achieve efficient generalisation. For instance, they can learn how variations in different users' insulin sensitivities impact the effectiveness of insulin injections in reducing blood glucose levels. In addition, to prevent overfitting, we use the model-ensemble policy optimisation (Section \ref{3.3}). The framework is shown in Figure \ref{fig1}.


\begin{figure*}[!ht]
    \centering
    \includegraphics[width=\linewidth]{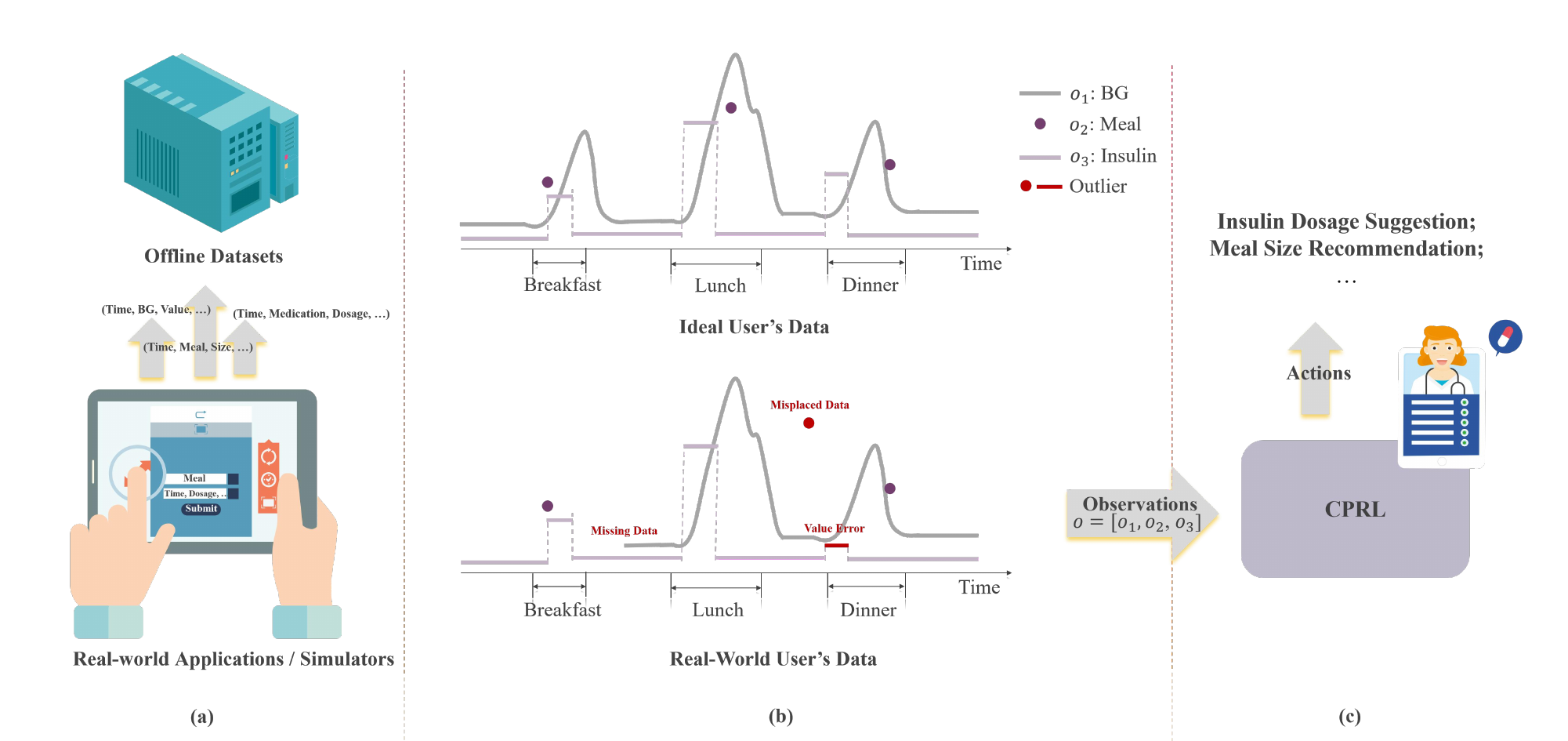}
    \caption{Problem Statement. (a) Source of the offline datasets. (b) Real-world offline datasets are highly suboptimal, encompassing missing data (e.g., omitted uploads), value errors (e.g., incorrect input values, miscalculated carbohydrate estimates), and misplaced data. Misplaced data pertains to data that is outside the anticipated time horizon. (c) Our CPRL utilises observations as input to furnish support for medical decision-making, including suggestions for insulin dosages and meal size recommendations.}
\label{exp_problem_statement}
\end{figure*}

\subsection{Dynamic Model: Hidden-Parameter Block Causal Prompting Dynamic \label{3.1}}

This section describes how to use the causal prompt to learn dynamic models - Hip-BCPDs. We first introduce how to design and apply causal prompts; secondly, we present the definition and learning methods of Hip-BCPDs.

\subsubsection{Causal Prompt}

Limited and sub-optimal offline data brings great difficulties to learning dynamic models. Learning in such resource-limited scenarios requires researchers to find ways to leverage existing experience, such as pre-trained models of similar tasks.

\textit{\textbf{Prompt}} is an approach to retrieving useful knowledge from pre-trained models. Similarly, we hope to utilise useful knowledge from existing dynamic models instead of training it from scratch. With prompt learning on the causal template, the learning task of dynamic models has been simplified to a masked language modelling problem wherein only a small fraction of all parameters need to be learned, similar to a fill-in-the-blanks task \citep{shin2020autoprompt}.

To construct the fill-in-the-blanks, a common structure applied to all tasks needs to be found and used as a template. A common structure exists among dynamic models of different tasks when the modelling method can be used for all the tasks \citep{zhang2020learning} inspired by causal learning. A causal relationship of variables has invariance in out-of-distribution generalisation. We exploit a causal graph to represent the common structure between different tasks. Specifically, we use the causal relationship in the existing model as the prompt, called \textit{\textbf{causal prompt}}, and quickly adapt the dynamic model to the current task with the help of offline data. Further, causal prompts are also used as prior knowledge to recover a robust environmental state. We explore the setting where the state space is latent and we have access to only high-dimensional observations. Causal prompt does not introduce large amounts of additional parameters in the above process \citep{shin2020autoprompt,qiu2020pre}.

\begin{table*}[!ht]
    \caption{\label{prompt-table} An example of a causal prompt from the glucose-insulin system.}
    \begin{tabular}{p{8cm}|p{7.7cm}} \toprule
    \hline
    \multicolumn{1}{p{8cm}|}{\textbf{Causal knowledge}} & \multicolumn{1}{p{7.7cm}}{\textbf{Causal prompt}} \\ \hline
    \multicolumn{1}{p{8cm}|}{ Insulin $I$ flows through liver. A portion $I_l$ is broken down and used by the liver; another $I_p$ flows into glucose. } & \multirow{3}{9cm}{Causal Knowledge $\mathcal{K}$ between 3 dimensions in state $ s = [s_1,s_2,s_3]$:\\ $\dot{s_2}=-(\theta_1 + \theta_3)\cdot s_2 + \theta_2 \cdot s_3 + s_1$; \\ $\dot{s_3}=-(\theta_2+\theta_4)\cdot s_4 + \theta_1 \cdot s_2$. } \\ \cline{1-1}
    \multicolumn{1}{p{8cm}|}{Insulin utilization by the liver: $\dot{I_l}=-(m_1+m_3)\cdot I_l + m_2\cdot I_p + I$. }  &   \\ \cline{1-1}
    \multicolumn{1}{p{8cm}|}{Plasma insulin: $\dot{I_p}=-(m_2+m_4)\cdot I_p + m_1\cdot I_l$. } &                                             \\ \hline
\bottomrule
    \multicolumn{2}{p{15cm}}{$\ast m_1,m_2,m_3,m_4$ are patient-specific parameters; $\dot{\ }$ is the value of a physical quantity at the next moment. }                                  \\ \hline
\bottomrule
\end{tabular}
\vspace{-10pt}
\end{table*}

Pre-trained causal knowledge is used as template to transfer the original task observations $o$ to the reconstructed observations $o^{prompt}$. The original observations $o$ (e.g., the total insulin \textcolor{myred}{$o$} in Figure \ref{fig1}) can generate a prompt $\mathcal{K}$ (e.g., ``\textcolor{myper}{Insulin broken down by the liver} \textcolor{myyellow}{$o^{prompt\ 1}$} \textcolor{myper}{and insulin flows into glucose} \textcolor{myyellow}{$o^{prompt\ 2}$} \textcolor{myper}{accounts for} \textcolor{mygreen}{$[MASK_1]$} \textcolor{myper}{\% and} \textcolor{mygreen}{$[MASK_2]$} \textcolor{myper}{\% of total insulin} \textcolor{myred}{$o$} \textcolor{myper}{respectively.}'') prompted by causal knowledge. The causal knowledge is about the position of input $o$, $[MASK]$, and $o^{prompt}$ in the causal prompt $\mathcal{K}$, where $[MASK]$ needs to be filled in by a masked model. In Figure \ref{fig1}, $[MASK]$ represents the utilisation of insulin by different patients. A masked model $p([MASK]|o, \mathcal{K})$ generate a unique distribution of $[MASK]$ according to the offline data of a patient. $[MASK]$ is formulated as $\theta$, which is the hidden parameter of the different patients mentioned above. The reconstructed state is represented as $s = [o, o^{prompt}]$ (shown as a graph structure in Figure \ref{fig2}), which will be used in downstream tasks.

In multi-task settings, dynamics vary across tasks, but the causal prompt $\mathcal{K}$ is shared. Then we show how to model this setting as Hip-BCPDs, where the changes in dynamics can be defined by hidden parameters (i.e., $\theta$), unifying dynamics across tasks as a single global function.

\subsubsection{Hip-BCPD}

The Hip-BCPD (illustrated in Figure \ref{fig2}(a)) is derived from the Hip-BMDP method \citep{zhang2020learning} (depicted in Figure \ref{fig2}(b)) and the CGD \citep{arrighi2012causal}. The Hip-BCPD assumes a shared causal graph amongst state variables across all environments and varying hidden parameters ($\theta$ in Figure \ref{fig2}(a)) among different environments. This assumption is pertinent to a range of real-world scenarios. For instance, within clinical medical settings, diverse patient groups have different responses to the same medical treatment, notwithstanding the desired outcome is the same. This assumption allows us to  provide accurate zero-shot generalisation to novel environments.

Formally, a Hip-BCPD family is described as a tuple $\left \langle \mathcal{S}, \mathcal{A},\mathcal{O}, \mathcal{R},\Theta, P_\Theta, T_\theta, q_{\theta}, \gamma,\mathcal{K} \right \rangle $ (visualised as a graphical model in Figure \ref{fig2}(a)), where $\mathcal{S}$ signifies a state space, $\mathcal{A}$ denotes an action space, $\mathcal{O}$ represents an observation space, $\mathcal{R}:\mathcal{S} \times \mathcal{A} \to \mathbb{R}$, $\Theta$ is the hidden-parameter space, $T_\theta(s_{t+1}|s_t)$ delineates the transition distribution for a specific task indicated by task parameter $\theta \backsim P_\Theta$, the emission function $q_{\theta}(o_t|s_t)$ maps states $s_t \in \mathcal{S}$ to observations $o_t \in \mathcal{O}$, $\gamma \in [0,1]$ is the discount factor, and $\mathcal{K}$ is the causal prompt shared across the family. We are assigned a label $j \in \{1, ..., N\}$ for each of $N$ environments. A masked model $P_{\Theta}$ is utilised to learn a candidate $\theta$ that unifies the transition dynamics $T_\theta$ and emission function $q_{\theta}$ across all environments, illustrated via a green box in Figure \ref{fig1}. As previously stated, $\theta$ is the formulaic representation of $[MASK]$ and state $s_{t+1}$ at time step $t+1$ can be reconstructed as:
\begin{equation}
    s_{t+1} = \left[ o_{t+1}, o^{prompt}_{t+1} \right] = \mathbb{E}_{\theta \backsim P_{\Theta}(o_t, \mathcal{K}) } T_{\theta}(o^{j}_t,a^{j}_t,\mathcal{K}), 
    \label{eq0}
\end{equation}

where the reconstructed state $s$ will be utilised in downstream tasks. A dynamic function $T_{\theta}(o^{j}_t,a^{j}_t,\mathcal{K})$ is employed to predict the state at the subsequent time step. The objective function at time step $t$ is formalised as:
\begin{equation}
\mathcal{L}(\zeta)=\sum_{j=1}^{N}\mathbb{E}_{\theta \backsim P_{\Theta}(o_t, \mathcal{K}) } \text{MSE}(q(\hat{o}^j_{t+1}|T_{\theta}(o^{j}_t,a^{j}_t,\mathcal{K})),o^{j}_{t+1}),
\label{eq1}\\
\end{equation}
where $\zeta$ symbolises all learnable parameters within Hip-BCPDs. $a_t$ symbolises the action executed by the downstream policy. It is noteworthy that each hidden parameter $\theta \backsim P_{\Theta}(o_t, \mathcal{K})$ epitomises the traits of a specific environment (i.e., user), such as the encoding of the user's insulin sensitivity factor within the hidden parameter. This consequently informs the insulin dosage in downstream insulin infusion tasks, thereby facilitating the downstream tasks to attain a higher average return and generalisation across diverse environments (further details regarding the policy gradient iteration process are elucidated in Section \ref{3.2}).

\subsection{Policy: Skill-Reuse Strategy  \label{3.2}}

Just as humans acquire, reuse, and build upon existing skills to tackle more complex tasks, agents should similarly have the capacity to learn and develop skills continually, hierarchically, and incrementally over time \citep{parisi2019continual}. To cultivate a policy imbued with a skill-reuse strategy, we employ a hierarchical policy architecture - the CCM \citep{yu2022causal} algorithm, for policy training. Leveraging the topology of graph structures, we partition a Hip-BCPD into $m$ subsystems, each representing a fraction of the entire Hip-BCPD. The agent masters skills within each subsystem owing to their internal independence and reutilises these learned skills via the cascade relationship interlinking the subsystems. Within the hierarchical RL framework, the high-level policy orchestrates division and reuse, whereas the low-level policy facilitates skill acquisition.

\textit{\textbf{High-level policy}} $\pi^{\text{\text{high}}}$ endeavours to segment a Hip-BCPD into $m$ subsystems and repurposes the policy of each subsystem. Throughout the segmentation process, based on the reconstructed state $s_t$ of a Hip-BCPD, the high-level policy dispatches the current division action $a^{\text{\text{high}}}$ and subsequently generates a subsystem for the low-level policy $\pi^{\text{\text{low}}}$. Following $C$ timesteps, $\pi^{\text{\text{high}}}$ assimilates outcomes from $\pi^{\text{\text{low}}}$ and formulates the ensuing action. In the course of this assimilation, the high-level policy reutilises learned low-level policies (i.e., skills) in light of the information flux amongst them. The reward function for the high-level policy $\pi^{\text{\text{high}}}$ is expressed as:
\begin{equation}
R^{\text{\text{high}}}t = \frac{1}{C}\sum{i=1}^{C} \gamma^i R^{\text{\text{low}}}(s_{t+i}),
\label{E1}
\end{equation}
where $\gamma$ denotes the discount factor, and $R^{\text{\text{low}}}(s_{t+i})$ signifies the single-step reward of the low-level policy $\pi^{\text{\text{low}}}$ when in state $s_{t+i}$.

\textit{\textbf{Low-level policy}} $\pi^{\text{\text{low}}}$, upon receiving an action $a^{\text{\text{high}}}$ from the high-level policy, needs to accomplish sub-tasks in accordance with high-level action $a^{\text{\text{high}}}$. It governs the subsystem based on state $s_{i+c}$ ($c \in [0, C]$). The reward function for the low-level policy is delineated as:
\begin{equation}
    R^{\text{\text{low}}}_t = \sum_{j=1}^{N}||T(s_t,a_t^{\text{\text{low}}},\mathcal{K},P_{\Theta})-q||^2_2 ,
    \label{E2}
\end{equation}
where $q$ epitomises the goal state for the subsystem control task, and $a_t^{\text{\text{low}}}$ is the action executed by $\pi^{\text{\text{low}}}$ at timestep $t$, evaluated by measuring the divergence between the subsequent state $s_{t+1} \sim T(s_t,a_t^{\text{\text{low}}},\mathcal{K},P_{\Theta})$ and the goal $q$.

\textbf{Optimisation and Training}. The design of objective functions is analogous to that in CCM \citep{yu2022causal}. At the high level, the objective function comprises two segments corresponding to division and reuse:
\begin{equation}
    \mathcal{L}(\epsilon) = \mathbb{E}\sum_{i=0}^{C}\gamma^i R^{\text{high}}(s_{t+i})/C - \mathcal{L}^{FCR},
    \label{E3}
\end{equation}
where $T$ indicates the maximal high-level episode duration. $\mathcal{L}^{\text{FCR}}$ represents the objective function for a forward coupled reasoning (FCR) module employed in the skills reuse process, elaborated in detail by \citep{yu2022causal}.
Low-level policies pertain to skill-learning scenarios across different subsystems. The objective function for a low-level policy mirrors whether the corresponding skill has been assimilated and is articulated as:
\begin{equation}
    \mathcal{L}(\phi) = \mathbb{E}\sum_{i=0}^{C}\gamma^i R^{\text{low}}(s_{t+i})/C,
    \label{E4}
\end{equation}
where $C$ is the maximal low-level episode length.

\subsection{Preventing Overfitting: Model Ensemble \label{3.3}}
In the process of learning both the dynamic model and the policy, there is a propensity for the learnt policy to exploit regions where data from the limited offline dataset is available, leading to instability during training. To mitigate this challenge, we employ a model ensemble strategy \citep{kurutach2018model} to maintain model uncertainty and regularise the learning trajectory.

First, we construct a set of Hip-BCPDs $\{ T_{\psi_1}, ... ,T_{\psi_M} \}$ (referred to as a \textit{model ensemble}) utilising the identical real-world dataset $\mathcal{D}$. These Hip-BCPDs undergo training through standard supervised learning methodologies as delineated in Section \ref{3.1}. Each Hip-BCPD is characterised by distinct initial weights and mini-batches randomly selected.

Second, the efficacy of the policy is evaluated by utilising the $M$ trained Hip-BCPDs. The training process is sustained provided a specific ratio surpasses a predetermined threshold. The proportion of models in which the policy demonstrates enhancement is computed as:
\begin{equation}
\frac{1}{M}\sum_{m=1}^{M} \mathbbm{1}[\hat{\eta}(\epsilon_{\text{new}};\phi_{\text{new}};T_{\psi_m})>\hat{\eta}(\epsilon_{\text{old}};\phi_{\text{old}};T_{\psi_m})],
\label{E5}
\end{equation}
where $\hat{\eta}$ signifies the performance of the dynamic model as articulated in Equation \ref{eq1}, $\epsilon$ and $\phi$ denote the learnable parameters within the policy $\pi$. A threshold of 70\% is utilised. Should the ratio fall below 70\% across five gradient updates (minor improvements are considered tolerable should performance augment), the current iteration will be concluded.

\section{Experiments}

The CPRL framework performs well at addressing decision-support challenges within suboptimal offline datasets, demonstrating potential applicability to a spectrum of downstream tasks with varying application contexts. CPRL is particularly tailored to address the diverse scenarios that require engagement with suboptimal datasets, such as missing data, value errors, and misplaced data, as depicted in Figure \ref{exp_problem_statement}.

Our offline dataset originates from the medical domain, offering recommendations for medical treatment decision-making to patients. Specifically, CPRL can assess each patient's physiological states (e.g., blood glucose levels, meal sizes, previous insulin dosages) and suggest future insulin dosage recommendations (illustrated in Figure \ref{exp_problem_statement}(c)).  As shown in Figure \ref{exp_problem_statement}(a),  the offline data include both simulated and real-world environments. This is to demonstrate the tolerance of various algorithms to noise, with the real-world offline dataset being highly suboptimal,  predominantly attributed to diminished adherence and the lack of standardisation in data uploaded by users, as demonstrated in Figure \ref{exp_problem_statement} (b). Noisy data encompasses missing data (e.g., omitted uploads), value error (e.g., erroneous input values, flawed carbohydrate assessments), and misplaced data, among other issues, unlike the simulated offline datasets, which do not contain such suboptimal data. Specifically, in Section \ref{4.4.5}, we provide detailed information about the data, showing that the offline dataset from \textit{Dnurse} APP for each user contains varying levels of noise. In Section \ref{4.1}, we first describe the generalisation of causal prompts and an awesome existing pre-trained model which are used to generate prompt templates. Subsequently, we describe the experimental setup in simulation and real-world experiments. We also illustrate the settings of the baseline methods. In Section \ref{4.4}, we compare our CPRL and baselines in the above two datasets and verify the effectiveness of causal prompt on model generalisation from different perspectives. Finally, detailed ablation studies and empirical analyses of robustness and generalisation are also reported. Datasets and codes for our CPRL will be available online. 

\begin{algorithm}[!ht]
	\caption{CPRL: Train Loop (Online Setting)} 
	\label{alg1} 
	\begin{algorithmic}[1]
        \STATE Initialize a policy $\pi = \{ \pi^{\text{high}}_{\epsilon},\pi^{\text{low}}_{\phi} \}$ and all Hip-BCPDs $T_{\psi_1},...,T_{\psi_m}$.
        \STATE Initialize an empty dataset $\mathcal{D}$.
		\REPEAT 
            \STATE Collect samples from the \textit{glucose-insulin system} $T$ using $\pi$ and add them to $\mathcal{D}$.
            \STATE Train all models using $\mathcal{D}$.
            \REPEAT
               \STATE Collect fictitious samples from $\{T_{\psi_i}\}_{i=1}^m$ using $\pi$.
               \STATE Update policy on fictitious samples.
               \STATE Estimate the performance $\hat{\eta}(\epsilon;\phi;\psi_i)$ for $i=1,..,m$.
            \UNTIL{the performances stop improving.}
        \STATE Optimize $\pi$ using all Hip-BCPDs.
		\UNTIL{the policy performs well in the \textit{glucose-insulin system} $T$.}
	\end{algorithmic} 
\end{algorithm}

\begin{algorithm}[!ht]
	\caption{CPRL: Train Loop (Offline Setting)} 
	\label{alg2} 
	\begin{algorithmic}[1]
        \STATE Initialize a policy $\pi = \{ \pi^{\text{high}}_{\epsilon},\pi^{\text{low}}_{\phi} \}$ and all Hip-BCPDs $T_{\psi_1},...,T_{\psi_m}$.
        \STATE Initialize an dataset $\mathcal{D}$ from real world.
		\REPEAT 
            \STATE Sample $m$ mini-batch $\{\mathcal{D}_j\}_{j=1}^{m}$ from $\mathcal{D}$.
            \STATE Train model $T_{\psi_j}$ using $\mathcal{D}_j$ for $j=1,...,m$.
            \REPEAT
               \STATE Collect fictitious samples from $\{T_{\psi_i}\}_{i=1}^m$ using $\pi$.
               \STATE Update policy on fictitious samples.
               \STATE Estimate the performance $\hat{\eta}(\epsilon;\phi;\psi_i)$ for $i=1,..,m$.
            \UNTIL{the performances stop improving.}
        \STATE Optimize $\pi$ using all Hip-BCPDs.
		\UNTIL{the policy performs well in real environment $T$.}
	\end{algorithmic} 
\end{algorithm}

\subsection{Design of Causal Prompt \label{4.1}}

We derived the causal prompt from the publicly accessible \textit{glucose-insulin system} \citep{dalla2007meal}. The \textit{glucose-insulin system} represents a dynamic model elucidating the process of glucose ingestion and absorption, developed through a complex triple tracer meal protocol to monitor glucose conversion dynamics within the meals of 204 healthy individuals. Additionally, the \textit{glucose-insulin system} forms the open-source portion of the \textit{DMMS.R} and \textit{T1DM} simulators, crafted by \textit{The Epsilon Group} and certified by the US Food and Drug Administration. These simulators are regularly employed by researchers to conduct simulation studies to assess treatment protocols or dosing algorithms before the launch of extensive clinical trials \citep{lee2020toward,zhu2020basal, sun2018dual, daskalaki2016model, li2019glunet}. Given the extensive data requirements for developing the \textit{glucose-insulin system}, which might not be attainable in resource-limited situations, we need to harness causal prompts for decision-making tasks in online applications. This methodology aims to guide and refine our dynamic model, ensuring its alignment with established physiological mechanisms.

The \textit{glucose-insulin system} comprises thirty dynamic human models concerning glucose ingestion and absorption, integrated via a shared graphical structure. Within this framework, each edge signifies a mathematical causal relationship between two variables, denoting a unit of causal knowledge. We adopt this graphical structure as our causal prompt, encompassing 35 parameters and 37 pieces of causal knowledge. An example provided in Table \ref{prompt-table} explicates three distinct pieces of causal knowledge sourced from the \textit{glucose-insulin system}, showcasing their application in creating a causal prompt. As outlined by Formula \ref{eq0}, $s_1=o$ corresponds to the environmental observation, whereas $s_2$ and $s_3$ are reconstructed observations (i.e., $o^{prompt}$) formulated in accordance with the causal prompt. The masked model $P_{\Theta}$ engages in the dynamic learning challenge by assimilating the hidden parameter vector $\theta=[\theta_1, \theta_2, \theta_3, \theta_4]$, encapsulating all modifiable parameters within both $T_{\theta}$, the transition dynamics, and $q_{\theta}(o|s)$, the observation model. This coherent approach facilitates the optimisation and comprehension of the system's dynamics.

\subsection{Experimental Environment Setups \label{4.2}}

\textbf{Simulation-based experiments.} We conduct simulation-based experiments within the \textit{glucose-insulin system}. In these experiments, involving 10 different adults, the agent's objective is to administer insulin dosages aimed at maintaining the patient's blood glucose level within the normative range (i.e. 70-180 mg/dL). Each step represents one minute, with the maximum duration of an episode being 1440 minutes, corresponding to one natural day.
To approximate real-world conditions, we introduce random noise into the simulation-based environment, thereby enhancing the agent's potential for generalisation to real-world scenarios. Specifically, each patient consumes three meals daily in the morning, noon, and evening, with meal sizes determined by carbohydrate content. The carbohydrate amount is calculated based on the patient's individual settings (e.g. correction factor, insulin-to-carbohydrate ratio), and a uniform distribution of random noise ranging from -10\% to +10\% in carbohydrate amount is applied. This random noise imitates uncertainties encountered in real-life scenarios (e.g. sensor inaccuracies and irregular eating patterns). In the simulation-based experiments, our proposed CPRL is deployed in an online setting, as detailed in Algorithm \ref{alg1}.

\textbf{Real-world experiments.} This experiment leverages offline data provided by the \textit{Dnurse} APP \textsuperscript{\ref {dnurse_web}}, a comprehensive online system dedicated to the management of diabetes. The offline dataset is notably varied and sub-optimal, comprising users' self-reported health and lifestyle data. We utilise data from nine Type-1 diabetes patients and apply denoising data processing techniques. We filtered the data for patients' meals, insulin dosages, blood glucose fluctuations, and insulin dose information, amalgamating them into a quadripartite data representation typical in RL - $(o, a, o',r)$, where observation $o=[o_1,o_2,o_3]$ includes the current blood glucose value $o_1$, meal size $o_2$, and previous insulin dosage $o_3$, with action $a$ representing the insulin dose. In real-world experiments, CPRL adopts offline learning, as elucidated in Algorithm \ref{alg2}.

\begin{center}
    \begin{table}
        \caption{\label{table:1} Results for simulation-based experiments. Each number is the normalised single-step average reward of the policy at the online testing, averaged over 6 random seeds. The negative rewards suggest that the policy fails to finish tasks. The names of environments are shown in the leftmost column. We bold the highest mean.}
            \centering
            \setlength{\tabcolsep}{2mm}{
            \begin{tabular}{cccccc}
                \toprule
                 & CPRL (ours) & DT & MOPO & MBPO & SAC\\
                \midrule
                Adult\#001 & \textbf{19.14} & -0.71 & -218.16 & -36.49 & -73.89 \\
                Adult\#002 & \textbf{16.28} & 2.34 & -218.16 & -82.52 & -12.17 \\
                Adult\#003 & \textbf{17.99} & -79.48 & -875.38 & -224.58 & -660.51 \\
                Adult\#004 & \textbf{20.27} & 8.08 & -125.76 & -291.33 & -394.82 \\
                Adult\#005 & \textbf{22.03} & -7.40 & -125.75 & -48.29 & -46.60 \\
                Adult\#006 & \textbf{20.63} & -325.67 & -1697.24 & -861.30 & -1000.18 \\
                Adult\#007 & \textbf{19.16} & 0.20 & -246.62 & -593.37 & -1023.52 \\
                Adult\#008 & \textbf{9.07} & 8.32 & -171.19 & 4.83 & -2.20 \\
                Adult\#009 & \textbf{15.09} & -2.01 & -301.80 & -61.27 & -93.79 \\
                Adult\#010 & \textbf{16.23} & -46.68 & -301.72 & -103.99 & -232.91 \\
               \midrule
                Average & \textbf{17.59} & -44.30 & -428.18 & -229.83 & -354.06 \\
                AVEDEV & \textbf{2.74} & 25.84 & 343.25 & 211.30 & 332.56 \\
                \bottomrule
            \end{tabular}}
    \end{table}
    \end{center}
    
    \begin{center}
    \begin{table}
        \caption{\label{table:2} Results for real-world experiments. Each number represents the normalized single-step average reward of the policy at the Dnurse offline dataset, averaged over 6 random seeds. User IDs are shown in the leftmost column. We bold the highest mean.}
        \centering
        \setlength{\tabcolsep}{2mm}{
        \begin{tabular}{cccccc}
            \toprule
             & CPRL (ours) & DT & MOPO & MBPO & SAC\\
            \midrule
            2366317 & \textbf{20.74} & -2307.84 & -125.30 & -1644.70 & -1867.80 \\
            3138156 & \textbf{14.45} & -662.25 & -130.47 & -1160.76 & -1045.54 \\
            100938 & \textbf{14.50} & -21.38 & -112.69 & -175.00 & -106.13 \\
            2007938 & \textbf{16.52} & -126.57 & -1150.65 & -217.40 & -762.81 \\
            2041527 & \textbf{21.08} & -131.29 & -132.78 & -261.20 & -361.60 \\
            2679594 & \textbf{9.29} & -36.26 & -9.62 & -22.49 & -74.98 \\
            2949314 & \textbf{16.36} & -160.92 & -1768.68 & -263.61 & -1087.27\\ 
            3198058 & \textbf{11.61} & -46.31 & -102.75 & -6.11 & -7.43 \\
            3261447 & \textbf{9.78} & 0.90 & -89.66 & -126.50 & -2.55 \\
           \midrule
            Average & \textbf{14.92} & -387.99 & -402.51 & -430.86 & -590.68 \\ 
            AVEDEV & \textbf{3.33} & 101.24 & 469.85 & 431.94 & 533.49  \\
            \bottomrule
        \end{tabular}}
    \end{table}
    \end{center}

Achieving effective control in both simulation and real-life conditions is formidable due to the intricate dynamics inherent in biological systems. The \textit{glucose-insulin system}, while built based on scaling real-world patient data, may not fully mimic real-world intricacies. Consequently, the distributional shift from simulation to real-world contexts presents a formidable challenge to the algorithm's generalisation capabilities. Moreover, the diverse dynamics of each patient necessitate a personalised approach to medical treatment.

\subsection{Baselines \label{4.3}}

As CPRL is a model-based offline RL framework, to show the performance of both dynamic model and policy respectively, we juxtapose it with several purely online or offline RL algorithms. Additionally, we compare it to notable methods that selectively integrate online or offline data into online or offline policy learning. These methods include:

\begin{figure*}
    \centering
    \includegraphics[width=\linewidth]{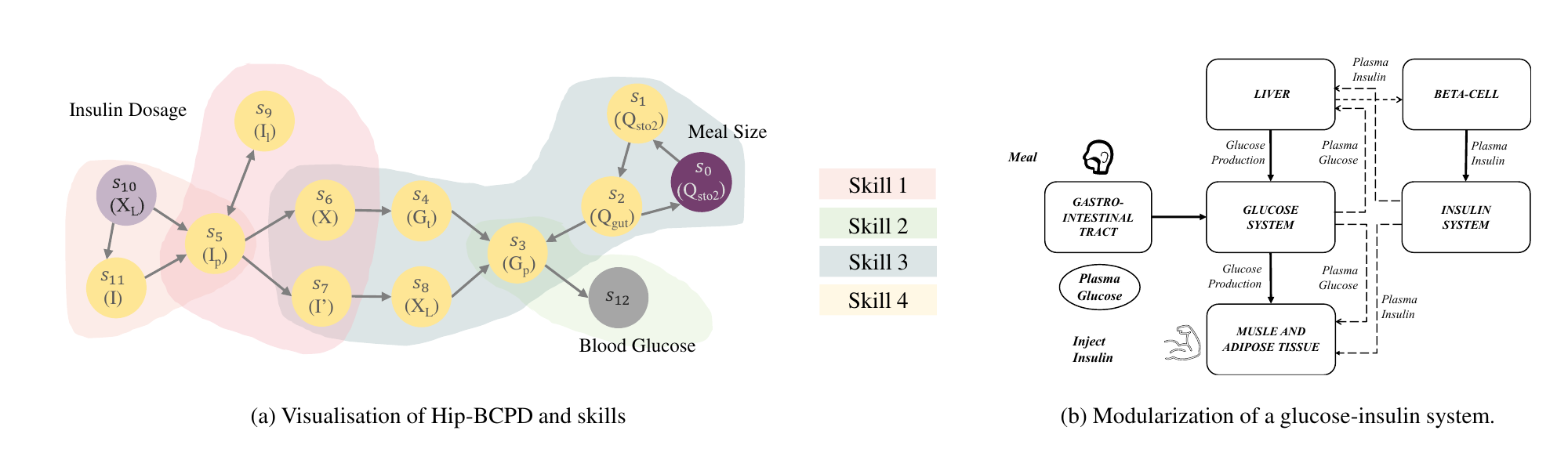}
    \caption{(a) The visualisation of learned Hip-BCPD and learned skills. (b) Modularisation of a glucose-insulin control system (cited from \citep{yu2022causal}). The glucose-insulin control system can be segmented into the insulin subsystem, glucose subsystem, and other unit process models, each necessitating mutual information and influence.}
    \label{visual}
\end{figure*}

\begin{itemize}
    \item SAC \citep{haarnoja2018soft}: the state-of-the-art online model-free RL algorithm, trained and evaluated within the simulator in the simulation-based experiment and optimal dynamic model (i.e., Hip-BCPDs) in the real-world experiment.
    \item MBPO \citep{janner2019trust}: an online model-based RL framework, based on SAC policy. MBPO is trained similarly to SAC. 
    \item MOPO \citep{yu2020mopo}: the state-of-the-art offline model-based RL algorithms, based on SAC policy. We train MOPO on the offline dataset sampled from the online training data of MBPO in the simulation-based experiment and the offline dataset from \textit{Dnurse} in the real-world experiment. It will suffer from diversity and distribution shifts in the offline dataset.
    \item Decision Transformer (DT) \citep{chen2021decision,zheng2022online}: a pioneering offline RL method, focusing on sequential decision-making through a transformer architecture. It enables the effective use of offline datasets without directly modelling environmental dynamics, beneficial in scenarios where accurate dynamic modelling is complex. DT offers a fresh approach to using suboptimal offline data for robust policy learning, and its comparison with CPRL promises to yield insightful distinctions.
\end{itemize}

\subsection{Comparative Evaluation of CPRL in Simulation and Real-World Experiments \label{4.4}}

\textbf{Simulation-based Experiments}

This experiment primarily seeks to validate the efficacy of CPRL, hence each method is trained across ten environments (i.e., ten adult patients) in scenarios not limited by resources. Specifically, MBPO and SAC are subjected to online training, entailing that data from the environment is sampled concurrently with the running of the algorithm. Conversely, CPRL, DT and MOPO undergo training in an offline modality, utilising previously amassed data.

Table \ref{table:1} delineates single-step average rewards across 1440 steps in an online assessment. Through juxtaposition with both online/offline and model-free/model-based RL models on simulation datasets, empirical investigations demonstrate that our proffered CPRL algorithm is capable of securing the most commendable single-step average rewards in the task of blood glucose management. The presence of negative average rewards in Table \ref{table:1} infers the inability of MOPO, MBPO, and SAC to satisfactorily complete tasks, whereas CPRL attains the most favourable positive rewards. Among all the baselines, DT, benefiting from its advanced model structure (Transformer), demonstrated superior blood glucose control performance; however, CPRL remains the state-of-the-art algorithm.

\textbf{\textit{CPRL's Superiority in Dynamic Model Reconstruction.}}
Upon comparing CPRL with MBPO (model-based), MOPO (model-based), and SAC (model-free), it becomes evident that model-based RL methodologies markedly elevate single-step average reward whilst mitigating variance. Further, we refer to DT as implicit model-based offline RL because it implicitly establishes a dynamic model. Specifically, we incorporate a prediction of the next state into the loss function. In a nutshell, CPRL's superiority over not only model-based RL (MBPO and MOPO) and implicit model-based RL (DT) underscores the effectiveness of Hip-BCPDs.

\textbf{\textit{CPRL's Efficacy in Handling Novel States.}}
In the training process of CPRL and MOPO, the offline dataset is sampled from the training transitions, collected when MBPO interacts with environments. Upon examining the outcomes of SAC, MBPO (online), MOPO (offline), and DT (offline), it becomes apparent that policy effectiveness markedly declines without the ability to generate additional transitions through environmental interactions. CPRL, however, outshines online RL models, indicating its ability to generalise from the offline dataset to out-of-distribution actions. Should there be shifts in the environmental dynamics, CPRL is adept at utilising the learned model for replanning.

\textbf{Real-world experiments} 

This phase introduces further constraints on the offline dataset; the suboptimal, diverse, and limited dataset from \textit{Dnurse} challenges the offline RL algorithms to prove robustness within highly suboptimal data contexts. In this experiment, CPRL, MOPO, and DT are trained using offline data from \textit{Dnurse}. As MBPO and SAC are predicated on online algorithms, they are trained using the optimal Hip-BCPDs in the online setting.

\begin{figure*}
    \centering
    \includegraphics[width=\linewidth]{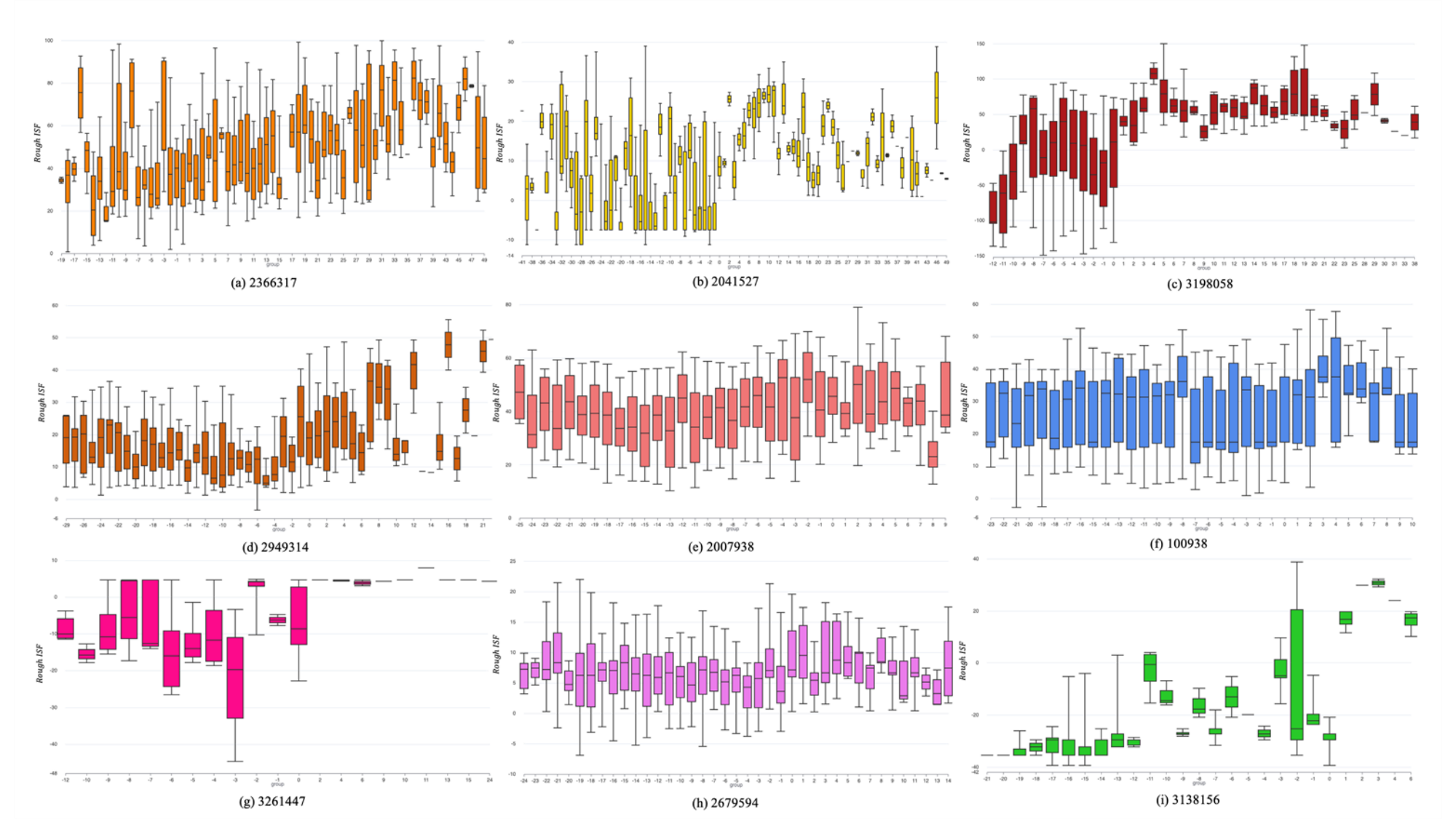}
    \caption{Box and Whisker Plots for different noise levels of the \textit{Dnurse} offline datasets.}
    \label{boxchart}
\end{figure*}

\textbf{\textit{CPRL's Adaptability to Limited, Highly Suboptimal, and Diverse Offline Datasets. }}
Foremost, the proposed dynamic modelling approach (i.e., Hip-BCPDs) proficiently recovers the environmental dynamics from suboptimal data, thereby eschewing the overfitting to noisy data. Learning trajectories for Hip-BCPDs, depicted in Figure \ref{fig3}(c), uniformly reach the lowest loss values (expanded upon in ablation studies). Additionally, CPRL's policies maintain robust performance even when confronted with out-of-distribution or noisy data. Table \ref{table:2} delineates the performance across four control methods. Except for CPRL, the performance of the other five baseline methods experiences severe degradation.

\subsection{Visualisation of Hip-BCPD and Interpretability of Learned Skills \label{4.6}}

We have visualised the Hip-BCPD structure for the glucose-insulin system and \textit{Dnurse} dataset as elucidated by the CPRL framework, depicted in Figure \ref{visual}(a). Each arrow in the figure symbolises a causal prompt, cumulatively amounting to 37. Every circle within the diagram denotes a dimension of the state, reconstructed by the emission function; where $s_0$, $s_{10}$, and $s_{12}$ equate to the observation $o$, and Formula (5) optimises the dynamic model $T$ by minimising the MSE distance between $s_0$, $s_{10}$, $s_{12}$, and $o$.

By visualising the skills indicated by the varied coloured shadows in Figure \ref{visual} (a), learned through the CPRL framework, we found that the learned skills can align with the glucose-insulin system. This visualisation aids in comprehending the interpretability and redundancy of skills. Owing to the Markov property manifested by the Hip-BCPD within the state space $\mathcal{S}$, and the boundedness as delineated in Definition 1, the scope of action effects exhibits a localised characteristic. For instance, meal dosage impacts only the blood sugar levels within the digestive system, with its effect waning with each increment in hops. \colorbox{myblue}{Skill 3} is linked to the ``gastrointestinal tract subsystem'' as shown in Figure \ref{visual}(b); \colorbox{mygreenshade}{Skill 2} encapsulates the ``insulin subsystem'', ``liver subsystem'', and ``muscle and adipose tissue subsystem'', signifying that insulin is accordingly allocated and utilised within the liver and the muscle and adipose tissue systems, wherein $I_p$ and $I_l$ (pmol/kg) denote insulin masses in plasma and liver, respectively; \colorbox{myredshade}{Skill 1} symbolises the transference of infused insulin from interstitial fluid to plasma; \colorbox{myyellowshade}{Skill 4} delineates the transference process of glucose from plasma to interstitial fluid.

\subsection{Robustness for different noisy and resource-limited versions of datasets \label{4.4.5}}

We roughly estimate the noisiness of user-uploaded meal size and insulin dosage data by calculating the Insulin Sensitivity Factor (ISF). In the scenario of insulin infusion, patients need to consider the carbohydrate content of their meal (i.e., meal size) and then inject the corresponding dose of insulin. The ISF reflects the power of a unit of insulin in the body and indicates how much one unit of rapid-acting insulin may drop a blood glucose level. Generally, one unit of rapid-acting insulin can metabolise 12-15 grams of carbohydrates. However, depending on each person's insulin sensitivity, it can vary from 6–30 grams or even more. Insulin sensitivity can differ among individuals, vary at different times, and be affected by the amount of physical activity and emotional stress. Hence, we can utilise the variation in a patient's ISF to ascertain the accuracy of their entered meal size and insulin dosage.

Since exercise data is provided in the \textit{Dnurse} data, when calculating the ISF using meal size, we actually use $meal\ size$ = $estimated\ carbohydrate\ content\ of\ meal$ - $estimated\ carbohydrate\ consumption\ of\ exercise$. Therefore, the ratio we calculate here is referred to as $Rough\ ISF$. $Rough\ ISF$ is only used for Box and Whisker Plots to visualise the noise level of offline datasets and is not used in the calculation process of CPRL.

In Figure \ref{boxchart}, ISFs for real-world \textit{Dnurse} offline datasets are analysed. In each patient's plot, each group contains 20 records. For each patient, the $Rough\ ISF$ should not be too large. Therefore, if the median between different groups varies significantly, or if the interquartile range within the same group is too large, it indicates that the data uploaded by the user is of a higher degree of sub-optimality.

The order from (a) to (i) in Figure \ref{boxchart} is sorted by the decreasing improvement brought by CPRL compared to baseline algorithms. We can observe that the median $Rough\ ISF$ of different groups for the three patients in the first row varies significantly and has a large interquartile range, for instance, user 3198058 varies from -100 to 100. The median $Rough\ ISF$ of patients in the last row does not change much. Although the median $Rough\ ISF$ difference for user 3138156 is not very small, the effect on model improvement is relatively limited due to the small amount of data (fewer groups). Our conclusions here are: 1) CPRL exhibits commendable robustness to highly optimal offline datasets; 2) CPRL possesses a degree of robustness in resource-limited scenarios, yet its performance significantly declines when the volume of data is too scant.

\subsection{Ablation Studies \label{4.5}}

\begin{figure*}
    \centering
    \includegraphics[width=1\linewidth]{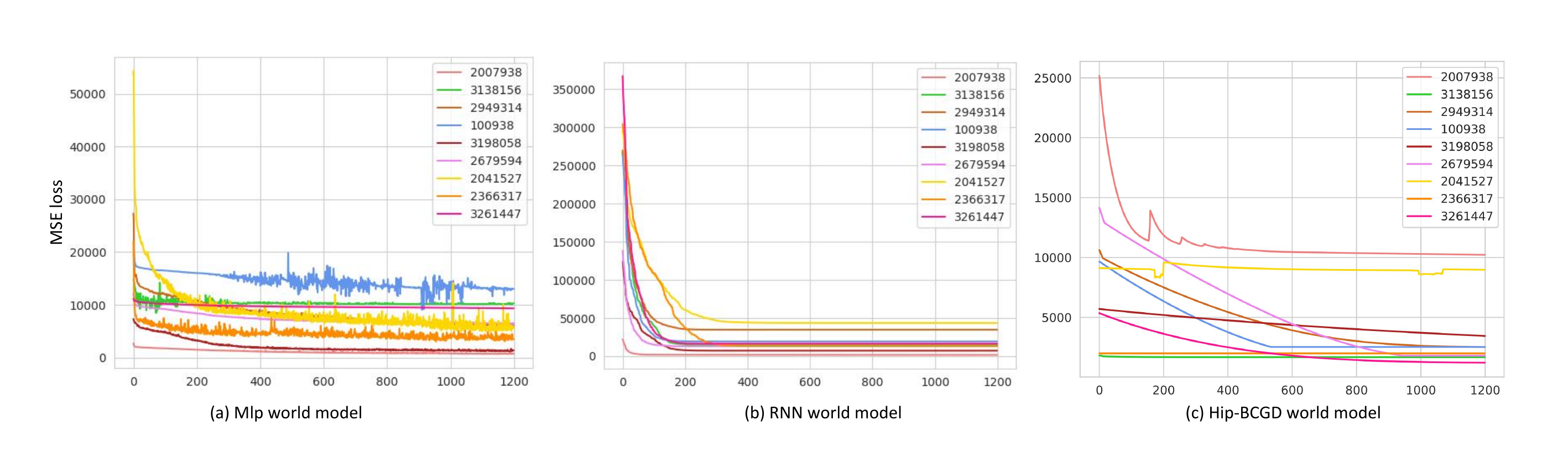}
    \caption{The learning curves of three dynamic models in \textit{Dnurse} offline datasets. The dynamic models and CCM are jointly trained in an offline setting. The labels at the top right of the figures indicate user IDs.}
    \label{fig3}
\end{figure*}
\begin{center}
    \begin{table}
        \caption{\label{table:3} Reults for ablation experiments. Comparisons are to CPRL (Hip-BCPD + CCM), MLP + CCM, and RNN + CCM. Each number is the normalized single-step average reward of the policy at the Dnurse offline dataset. User IDs are shown in the leftmost column. }
            \centering
            \setlength{\tabcolsep}{3.5mm}{
            \begin{tabular}{cccc}
                \toprule
                & CPRL & MLP+CCM & RNN+CCM \\
                \midrule
                2366317& \textbf{20.74}& -117.55& -117.51 \\
                3138156& \textbf{14.45}& -119.40& -119.80 \\
                100938& \textbf{14.50}& -120.14& -122.03 \\
                2007938& \textbf{16.52}& -127.81& -123.97 \\
                2041527& \textbf{21.08}& -121.34& -116.06 \\
                2679594& \textbf{9.29}& -123.30& -122.90 \\
                2949314& \textbf{16.36}& -125.36& -121.66 \\
                3198058& \textbf{11.61}& -127.64& -119.29 \\
                3261447& \textbf{9.78}& -119.50& -120.03 \\
               \midrule
                Average& \textbf{14.92}& -122.45& -120.36 \\ 
                \bottomrule
            \end{tabular}}
    \end{table}
    \end{center}

In this section, we examine the influence of two pivotal components of CPRL: 1) the dynamic modelling method - Hip-BCPD, and 2) the policy with a skill-reuse strategy - CCM. Employing the offline dataset from real-world experiments, we introduce two novel control methods — MLP+CCM and RNN+CCM, wherein the dynamic models are substituted (with MLP and RNN models, respectively), while the policies remain unchanged. The training processes are the same as the joint training protocol as delineated in Algorithm \ref{alg2}. The learning trajectories of the three dynamic models are depicted in Figure \ref{fig3}, with the single-step average rewards achieved by the control methods presented in Table \ref{table:3}.

\textbf{\textit{How much does Hip-BCPDs/causal prompting affect?}} By substituting Hip-BCPD with RNN and MLP, we can observe the effects of removing causal prompts and directly fitting offline datasets with different neural networks. The learning curves of Hip-BCPDs converge to the lowest errors gradually in Figure \ref{fig3}, while RNN and MLP still maintain significant errors. As shown in Table \ref{table:3}, the efficacy of control methods markedly diminishes subsequent to the dynamic models' substitution, failing to regulate the patient's blood glucose within the targeted range (i.e., accruing negative rewards). Consequently, two deductions can be made: 1) A superior dynamic model substantially bolsters the control approach, harbouring the capacity for enhanced sample efficiency in resource-constrained settings. 2) The causal prompting element of Hip-BCPD proves optimal for handling suboptimal data.

\textbf{\textit{And how much does policy with a skill-reuse strategy work?}}
The skill-reuse strategy aims to augment policy generalisation. The experiments reveal two key observations: 1) MLP+CCM and RNN+CCM (as shown in Table \ref{table:3}) consistently achieve higher single-step average rewards compared to MBPO, SAC, and MOPO (as indicated in Table \ref{table:2}), validating the significant role of the skill-reuse strategy, even in the absence of optimal dynamic models. 2) Control methods incorporating the CCM policy exhibit reduced variances across patients, notwithstanding the diverse dynamics of each patient. In conclusion, learning a single policy with common skills across different tasks, instead of training policies for each task separately, will avoid brittleness due to overspecialisation. When encountering a new environment, the single policy can reuse shared common skills to achieve stable generalisation, even if available datasets are limited or noisy.

\section{Conclusion}
In this study, we propose a novel framework, CPRL, to tackle model-based offline RL challenges. Experiments conducted within the real-world, large-scale online system, the \textit{Dnurse} APP, demonstrate that CPRL is capable of handling highly suboptimal and diverse offline datasets. The introduced Hip-BCPDs leverage a shared causal graph structure across tasks and differentiate tasks using hidden parameters. Evidence from our experiments suggests that Hip-BCPDs can effectively generalise to out-of-distribution data and bolster downstream policies. The dynamic model, informed by resource-limited scenarios, outperforms traditional online settings, indicating that insights derived from causal prompts in pre-trained models can address data scarcity and mitigate noise. Moreover, ablation studies affirm that the hierarchical policy with a skill-reuse strategy is adept at managing novel states. Theoretically and empirically validated, CPRL offers viable strategies for offline RL to facilitate online applications. Future work will explore the interpretability of learned skills and the implementation of online deployment.

\section*{Acknowledgments}

This study was supported in part by grants from the National Science and Technology Major Project [ZDYF20220008-02], the National Key R\&D Program of China [2021ZD0113302], and the National Natural Science Foundation of China [62006063]. Besides, the authors wish to acknowledge the \textit{Dnurse} Company for its medical side of resources and assistance.

\bibliographystyle{unsrtnat}
\bibliography{references}

\end{document}